\documentclass[letterpaper, 10 pt, conference]{ieeeconf} 
\IEEEoverridecommandlockouts
\usepackage{amsmath} 
\usepackage{cite}
\usepackage{graphicx}
\usepackage{url} 
\usepackage[hidelinks]{hyperref}
\usepackage{algorithm}
\usepackage{algpseudocode}
\usepackage{overpic}
\usepackage{array}
\usepackage{multirow}
\usepackage{booktabs}
\usepackage{amssymb} 

\title{\LARGE \bf
CompdVision: Combining Near-Field 3D Visual and Tactile Sensing \\ Using a Compact Compound-Eye Imaging System
}

\author{Lifan Luo, Boyang Zhang, Zhijie Peng, Yik Kin Cheung, Guanlan Zhang,  \\
Zhigang Li, Michael Yu Wang and Hongyu Yu
\thanks{Lifan Luo, Boyang Zhang, Zhijie Peng, Yik Kin Cheung, Zhigang Li are with the Department of Mechanical and Aerospace Engineering, The Hong Kong University of Science and Technology, Hong Kong, China. (e-mail: 
\href{mailto:lluoan@connect.ust.hk}{lluoan@connect.ust.hk}, 
\href{mailto:bzhangcd@connect.ust.hk}{bzhangcd@connect.ust.hk}, 
\href{mailto:zpengal@connect.ust.hk}{zpengal@connect.ust.hk}, 
\href{mailto:ykcheungab@connect.ust.hk}{ykcheungab@connect.ust.hk}, 
\href{mailto:mezli@ust.hk}{mezli@ust.hk})} 
\thanks{Guanlan Zhang is with the Department of Individualized Interdisciplinary Program (ROAS), The Hong Kong University of Science and Technology, Hong Kong, China. (e-mail:\href{mailto:gzhangaq@connect.ust.hk}{gzhangaq@connect.ust.hk})}
\thanks{Michael Yu Wang is with the School of Engineering, Great Bay University, Songshan Lake, Dongguan, Guangdong, China. (e-mail: 
\href{mailto:mywang@gbu.edu.cn}{mywang@gbu.edu.cn})}
\thanks{Hongyu Yu is with the Department of Mechanical and Aerospace Engineering, The Hong Kong University of Science and Technology, Hong Kong, China, and also with the HKUST Shenzhen-Hong Kong Collaborative Innovation Research Institute, Shenzhen 518000, China. (e-mail: 
\href{mailto:hongyuyu@ust.hk}{hongyuyu@ust.hk})}   
}         

\begin{document}

\maketitle
\thispagestyle{empty}
\pagestyle{empty}

\begin{abstract}
As automation technologies advance, the need for compact and multi-modal sensors in robotic applications is growing. To address this demand, we introduce CompdVision, a novel sensor that employs a compound-eye imaging system to combine near-field 3D visual and tactile sensing within a compact form factor.
CompdVision utilizes two types of vision units to address diverse sensing needs, eliminating the need for complex modality conversion. Stereo units with far-focus lenses can see through the transparent elastomer for depth estimation beyond the contact surface. Simultaneously, tactile units with near-focus lenses track the movement of markers embedded in the elastomer to obtain contact deformation.
Experimental results validate the sensor's superior performance in 3D visual and tactile sensing, proving its capability for reliable external object depth estimation and precise measurement of tangential and normal contact forces.
The dual modalities and compact design make the sensor a versatile tool for robotic manipulation.
Videos are available at \url{hkust-orisys.github.io/CompdVision}.
\end{abstract}

\section{Introduction}   

3D visual and tactile sensing significantly advance robotic manipulation by using depth cameras and tactile sensors, enhancing the robot's capabilities in dynamic environments \cite{visuotactile3}\cite{visuotactile4}. This fusion provides depth perception, object recognition and contact information, essential for executing robotic tasks with high autonomy and adaptability \cite{visuotactile1}\cite{visuotactile2}.

\begin{figure}[!t]
      \centering
      \begin{overpic}[width=\linewidth]{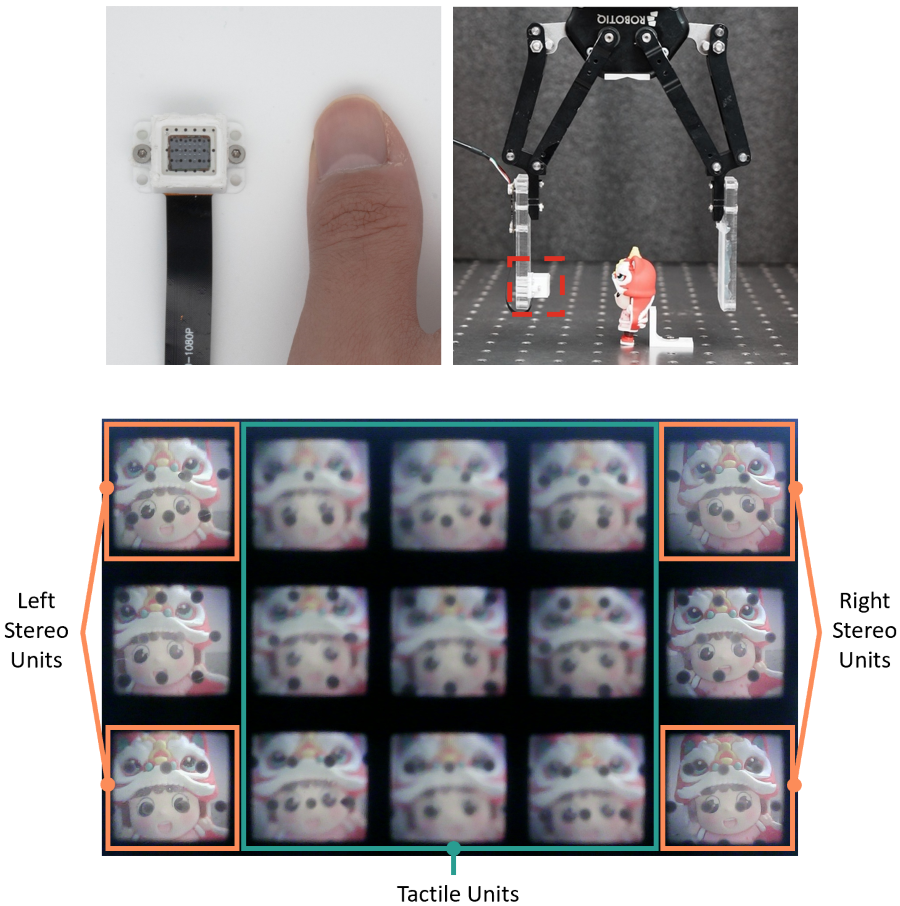}
      \put(30.5,55.5) {\small (a)}
      \put(67.5,55.5) {\small (b)}
      \put(47.5,-3.5) {\small (c)}
      \end{overpic}
      \vspace{-0.3cm}
      \caption{(a) Human thumb next to CompdVision. (b) CompdVision (red) is mounted on a gripper. (c) The compound-eye imaging system utilizes side stereo units with far-focus lenses for depth estimation, while the central tactile units, equipped with near-focus lenses, reduce external noise to enable precise tracking of marker movements for tactile sensing.}
      \label{fig:concept}
\vspace{-0.5cm}
\end{figure}

Vision-based tactile sensors, revolutionizing robotic touch perception by translating physical interactions into visual data \cite{gelsight2}, have proven invaluable across diverse applications \cite{application1}\cite{application2}. Miniaturization efforts, notably through the compound-eye imaging system \cite{compoundeye}, have achieved sensor sizes akin to human fingertips \cite{omnitact}\cite{pinhole}. Although focused on tactile sensing,  the system's design, where each vision unit operates independently and is customizable, paves the way for developing compact and multi-modal sensors.

Using cameras and tactile sensors separately has been a common approach in robotics. A single sensor combining visual and tactile sensing offers a significant advancement, improving robots' interaction and perception capabilities \cite{fingervision}. However, this innovation encounters some challenges. The reliance on 2D visual sensing limits depth perception and spatial understanding. Low-rate conversion between visual and tactile modalities \cite{spectac}\cite{tirgel}, presents challenges in dynamic manipulation tasks \cite{sts2}.
Furthermore, achieving a compact sensor combining 3D visual and tactile sensing is difficult due to bulky camera modules and the considerable spacing between stereo cameras. Such spacing results in non-overlapping fields of view when objects are close to cameras, necessitating the placement of the sensor surface centimeters away from cameras to perceive depth for close-up interactions \cite{Stereotac}. Addressing these challenges is vital for enabling precise and adaptable robotic manipulation.

\begin{table*}[!t]
\vspace{0.25cm}
\caption{Comparison of CompdVision and other Sensors}
\label{tab:comparison}
\centering
\begin{tabular}{|m{2.6cm}|m{1.8cm}|m{1.8cm}|m{2.5cm}|m{3cm}|m{2cm}|}
\hline
\multicolumn{1}{|c|}{\multirow{2}{*}{Sensor}} & \multicolumn{2}{c|}{Modality} & \multicolumn{1}{c|}{\multirow{2}{*}{Modality conversion}} & \multicolumn{1}{c|}{\multirow{2}{*}{Sensing rate}} & \multicolumn{1}{c|}{\multirow{2}{*}{Size (in mm)}} \\
\cline{2-3}
& \centering Vision & \centering Tactile & & & \\
\hline
GelSight\cite{gelsight2} & No & Force + Shape & — & 30fps & 35$\times$35$\times$60\\
\hline
OmniTact\cite{omnitact} & No & Shape & — & 30fps & $\varnothing$30$\times$33\\
\hline
Zhang et al. \cite{pinhole} & No & Force & — & 30fps & 14.5$\times$14.5$\times$5\\
\hline
FingerVision\cite{fingervision} & RGB, limited & Force & Simultaneous sensing & 30fps & 47$\times$40$\times$30 \\
\hline
Finger-STS\cite{sts2} & RGB & Marker Flow & LEDs & NA & 50$\times$50 \\
\hline
SpecTac\cite{spectac} & RGB & Force & \raggedright LEDs & 0.2s visual-tactile cycle & NA\\
\hline
StereoTac\cite{Stereotac} & RGBD & Shape & \raggedright LEDs & 4Hz LEDs alternating rate & NA\\
\hline
\hline
\textbf{CompdVision (Ours)} & RGBD & Force & Simultaneous sensing & 25fps & 22$\times$14$\times$14 \\
\hline
\end{tabular}
\vspace{-0.25cm}
\end{table*}

In this paper, we introduce CompdVision, a sensor that leverages a compound-eye imaging system to combine near-field 3D visual and tactile sensing within a compact form factor. Before making contact, it collects visual and depth information, and after contact, it offers deformation field and force measurements. The sensor's compact size, along with its ability to perform 3D visual and tactile sensing simultaneously without the need for modality conversion, advances the field of robotic sensing.

The CompdVision sensor is equipped with a 3$\times$5 grid of vision units, as shown in Fig. \ref{fig:concept}.
The vision units at the top-left and bottom-left corners constitute the left stereo pair, while the units at the top-right and bottom-right corners form the right stereo pair. Equipped with far-focus lenses, two pairs are dedicated to seeing through the transparent elastomer and capturing visual data, thereby facilitating robust depth estimation beyond the contact surface.
Simultaneously, the remaining nine vision units, centrally located on the grid, are designed to track the displacement of markers on the elastomer for tactile sensing. 
The partial views captured by each tactile unit are stitched, expanding the field of view. These units are equipped with near-focus lenses, which minimize noise interference from external scenes for enhanced tactile sensing capabilities. 
The placement of stereo vision units ensures that depth estimation effectively covers the central area where tactile sensing occurs. 

The contributions of this paper are as follows:

\begin{itemize}
\item  Employ a compound-eye imaging system for a compact and dual-modal sensor.

\item Achievement of 3D visual sensing with micro-scale vision units for near-field perception.

\item Simultaneous visual and tactile sensing using a dual-focus microlens array, eliminating modality conversion.
\end{itemize}

The rest of the paper is organized as follows: Section II provides an overview of related work; Section III details the sensor design and fabrication; Section IV presents the sensing principles of both the stereo and tactile units; Section V presents the experimental setup and corresponding results; and Section VI provides a conclusion and future work.

\section{Related work}
\subsection{Vision-Based Tactile Sensor}

In 2009, Johnson et al. \cite{gelsight1} introduced a tactile sensor using photometric stereo techniques and a tri-color lighting setup to reconstruct surface deformations. Yuan et al. \cite{gelsight2} enhanced this by adding dot markers for force estimation. The TacTip sensors \cite{tactip} used internal pins for deformation tracking. Lambeta et al. \cite{digit} developed Digit sensors for a robotic hand, enabling intricate object manipulation.

Despite this progress, the sensor size remained a challenge. The GelSlim sensor \cite{gelslim} employed light guides and mirrors for a slimmer profile. The OmniTact sensor \cite{omnitact} and Trueeb et al.'s design \cite{eye2} achieved compact dimensions with a multi-camera setup, pushing the boundaries of sensor miniaturization. 
However, these attempts faced challenges due to the inherent bulkiness of camera modules, even when employing fish-eye or micro-cameras.
Zhang et al. \cite{pinhole} innovated by integrating micro-scale vision units into a compound-eye imaging system, significantly slimming sensor thickness to 5 mm. Chen et al. \cite{lens} further optimized this with a microlens array, enabling high-resolution imaging without internal lighting. While these sensors with compound eyes were focused on tactile sensing, their innovative design laid the groundwork for developing compact and multi-modal sensors. Each vision unit within the system operates independently and can be customized for specific needs, enabling the combination of diverse modalities.

\subsection{Sensor Combining Visual and Tactile Sensing}

FingerVision \cite{fingervision} demonstrates visual-tactile sensing with its transparent skin and dot markers. However, its internal RGB camera's focus on markers for enhanced tracking restricts it to close object visualization. TIRgel \cite{tirgel} shows the conversion between tactile and visual modality via focus adjustment. Hogan et al. \cite{sts1} developed a sensor with semi-transparent skin that operates as either a tactile or visual sensor depending on internal lighting conditions. Such modality conversion limits it to switching to tactile sensing only after making contact \cite{sts2}.
SpecTac \cite{spectac} utilizes UV LEDs and fluorescent markers on a transparent elastomer for visual-tactile sensing. The sensor alternates between modalities by flickering the UV LEDs in a duty cycle. This method of modality conversion leads to a low visual-tactile sensing rate.

Single-camera setups in these sensors limit depth estimation capabilities, crucial for precise external object interactions. Shimonomura et al. \cite{eye1} introduced a sensor combining infrared and stereo cameras for tactile and depth sensing, respectively, though its rigid surface limits contact detail acquisition. 
StereoTac \cite{Stereotac}, featuring a semi-transparent skin, alternates between modes using LED lighting but faces challenges with low-rate modality conversion and struggles with compact integration due to the bulkiness of camera modules. Additionally, it shows a great challenge in designing compact sensors combining 3D visual and tactile sensing due to the millimeter-scale spacing between stereo cameras. This spacing prevents complete overlap of the cameras' fields of view, particularly resulting in blind spots and impaired depth perception when objects are close. Consequently, the sensor surface must be positioned centimeters away from the cameras for close-up interactions, restricting the development of compact sensors.
Furthermore, while some sensors can reconstruct the high-resolution shape of the sensor surface, they lack the capability for marker tracking or force estimation, essential for detecting slippage \cite{dong2017improved}.

In comparison to existing sensors (partially summarized in Table \ref{tab:comparison}), we develop a compact, dual-modal sensor using a compound-eye imaging system. We achieve near-field 3D visual sensing with micro-scale vision units and simultaneous visual and tactile sensing through a dual-focus microlens array without the need for modality conversion.

\section{Sensor Design and Fabrication}

\begin{figure}[!t]
\vspace{0.25cm}
      \centering
      \begin{overpic}[width=0.75\linewidth]{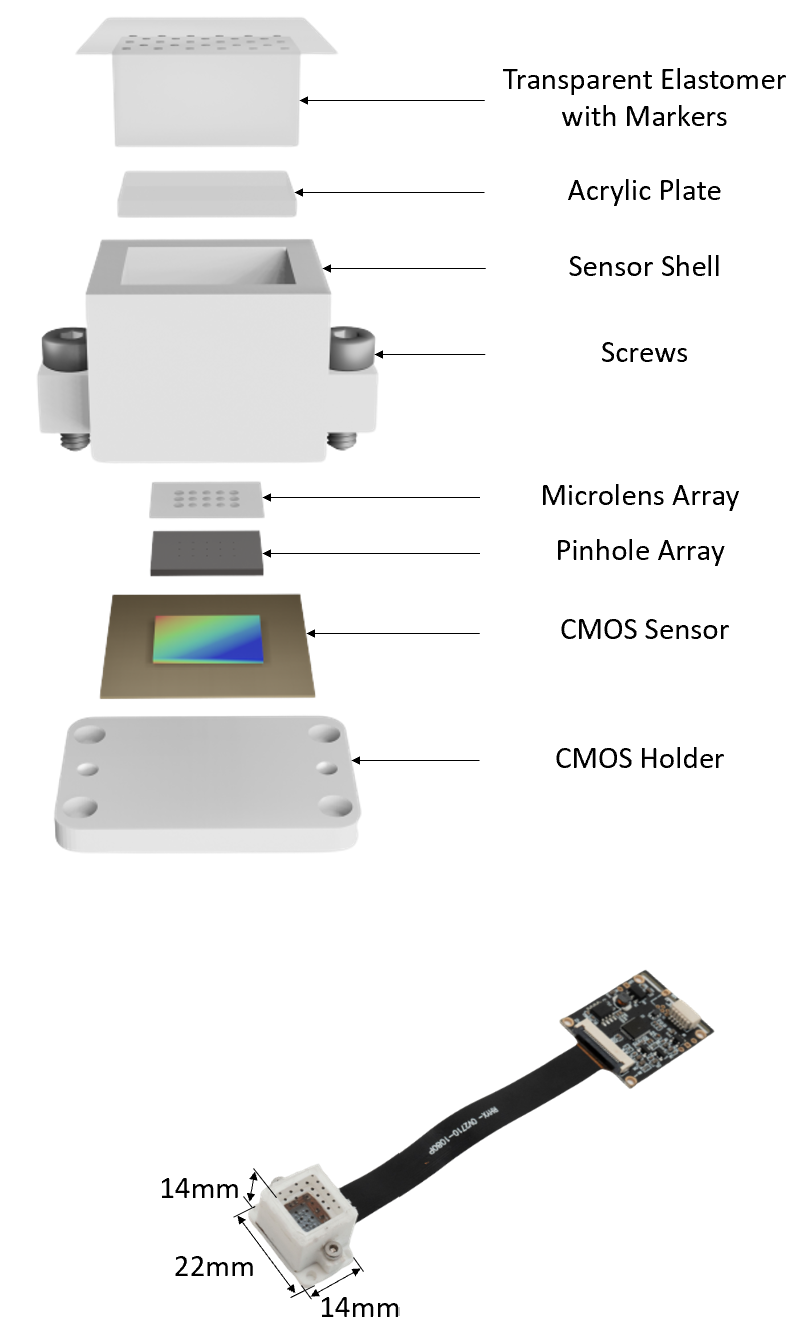}
      \put(27.5,32) {\small (a)}
      \put(27.5,-2) {\small (b)}
      \end{overpic}
      \caption{(a) Exploded view of the CompdVision sensor. (b) Prototype and dimensions of the sensor.}
      \label{fig:design}
\vspace{-0.25cm}
\end{figure}

The sensor comprises two subsystems: the touching subsystem and the imaging subsystem. An exploded view of the sensor is depicted in Fig. \ref{fig:design}(a), with the prototype and dimensions shown in Fig. \ref{fig:design}(b).

The touching subsystem, utilizing transparent elastomer with embedded dot markers, serves both structural and functional purposes, allowing for visual and tactile data collection. The transparency of the elastomer enables stereo units within the imaging subsystem to capture external visual data, while the tactile units track the markers' movements to obtain surface deformations. The touching subsystem protects the entire assembly and houses the imaging subsystem, which is crucial in processing both visual and tactile inputs. The compound-eye design of the imaging subsystem is customized to fulfill the different requirements on focal length for stereo and tactile vision units, ensuring the sensor's ability to capture and interpret 3D visual and tactile information.

The touching subsystem's fabrication begins by adhering an acrylic plate to the 3D-printed sensor shell using UV glue. 
Subsequently, Solaris (Smooth-On, Inc.), a transparent soft silicone, is poured into the structure, forming the base elastomer layer. Following the curing of Solaris, the sensing pattern is created by stamping silicone-based ink onto the base elastomer layer using a 3D-printed pillar array. The ink is composed of Psycho Paint, Smooth-On Silc-Pigment 9 black color paste, and silicone thinner mixed in a weight ratio of 8:6:3. A second layer of Solaris elastomer is then poured onto the mold and covered by the previous component, resulting in a protective elastomer layer that shields the dot markers. The total thickness of the elastomer is about 6mm.

The imaging subsystem comprises a CMOS sensor (OV 2710, OmniVision) and a vision unit array which is constructed by aligning a microlens array (MLA) with a pinhole array. The pinhole array is fabricated from a 525$\mu$m-thick silicon wafer, DRIE etched to form 100$\mu$m diameter circular holes at a 10$\mu$m depth and square holes of dimensions 500$\times$467 $\mu$m at a 515$\mu$m depth.
Fabricated using a lithographic process on a silicon wafer followed by reflow and molding \cite{lens}, the MLA possesses two distinct sizes, designed to meet the requirements of the stereo and tactile units. The stereo units demand far-focus lenses for external visual capture, while the tactile units use near-focus lenses for dot marker tracking and noise reduction. The final MLA features a sag height of 120$\mu$m, with lens diameters of 600$\mu$m for tactile units and 700$\mu$m for stereo units. 
The imaging subsystem is completed by affixing the vision unit array, formed by aligning the MLA with the pinhole array, onto the CMOS sensor. The spacing between adjacent vision units is 1mm.

\begin{figure*}[!t]
\vspace{0.25cm}
      \centering
      \begin{overpic}[width=\linewidth]{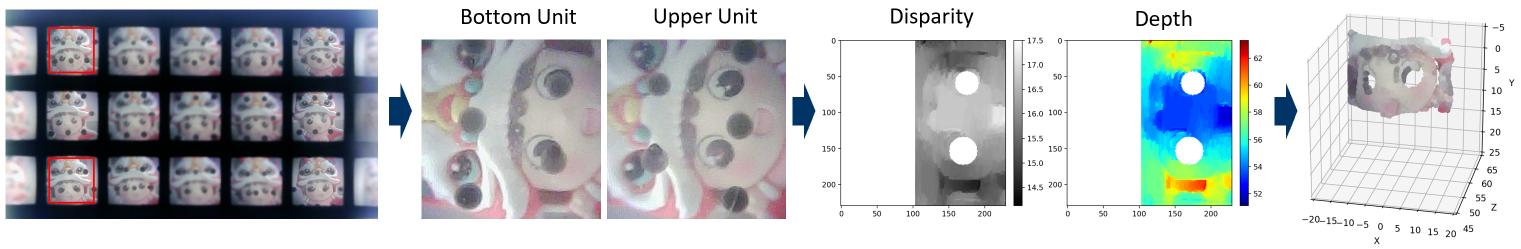}
      \put(12.5,-1.5) {\small (a)}
      \put(38.5,-1.5) {\small (b)}
      \put(67.5,-1.5) {\small (c)}
      \put(91,-1.5) {\small (d)}
      \end{overpic}
      \vspace{-0.3cm}
      \caption{Scheme of stereo depth estimation: (a) Extraction of image tiles of the left stereo units. (b) Transposition and rectification of image tiles of the bottom and upper units to achieve horizontal alignment. (c) Application of the SGBM algorithm to generate the disparity and depth maps, with marker areas removed. (d) Final 3D reconstruction result derived from the left stereo units.}
      \label{fig:stereo_process}
\end{figure*}

\begin{figure*}[!t]
      \centering
      \begin{overpic}[width=0.7\linewidth]{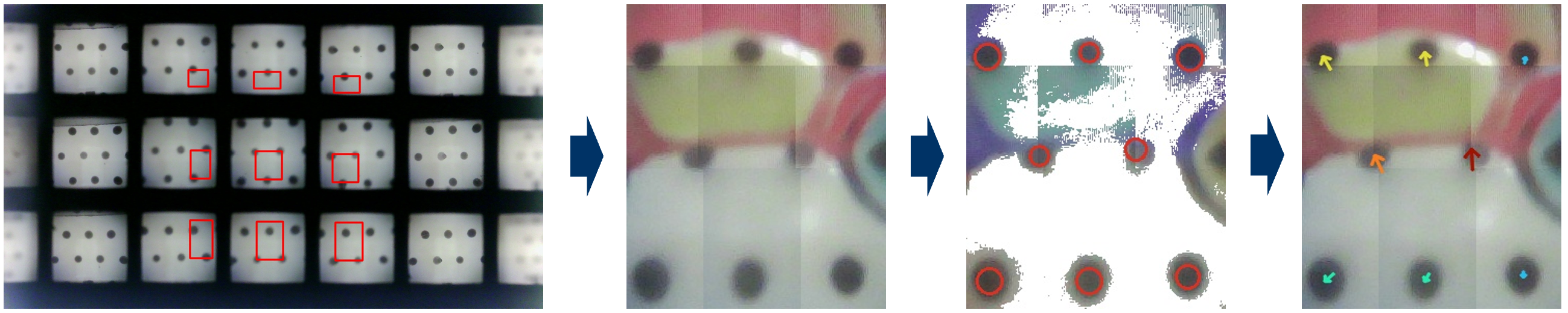}
      \put(16,-2) {\small (a)}
      \put(47,-2) {\small (b)}
      \put(68,-2) {\small (c)}
      \put(90,-2) {\small (d)}
      \end{overpic}
      \caption{Scheme of tactile sensing: (a) An initial image is used to crop ROIs based on the centers of common markers. These cropped ROIs are then stitched together in the non-overlapping areas to form a single image. (b) Utilizing the same cropping regions, a stitched image is formed after contact. (c) The RGB image is converted to HSV color space and thresholded to isolate the markers, followed by using SimpleBlobDetector for marker localization. (d) The detected blob points are compared and matched with those from the initial image, establishing correspondences based on nearest neighbors.}
      \label{fig:tactile_process}
      \vspace{-0.25cm}
\end{figure*}
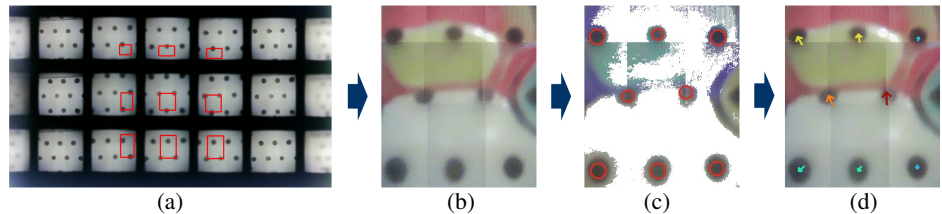

Finally, the touching subsystem and the imaging subsystem are jointly secured using screws, ensuring the stability and integrity of the assembled sensor.

\section{Sensing Principle}

This section details the sensing principles of the 3D visual and tactile modalities. The 3D visual modality enables depth perception through the sensor's surface, while the tactile modality captures surface deformation. We also present the simultaneous sensing of these two modalities.

\subsection{3D Visual Modality}

Stereo vision, a technique mimicking human vision, extracts 3D information from scenes captured from differing viewpoints \cite{stereo}. By positioning vision units at opposite corners, our sensor forms stereo pairs, ensuring sufficient overlap in their fields of view for effective stereo matching, especially when objects come into contact with the sensor. This configuration also guarantees high-resolution depth perception over extended distances. Furthermore, the placement of two pairs on the sides ensures coverage of the central area, where tactile sensing occurs.

Stereo calibration, conducted with a chessboard pattern using MATLAB's Stereo Camera Calibrator \cite{matlab}, emphasizes our sensor's small baseline (the distance between vision units), crucial for enabling near-field 3D visual sensing capabilities. Unlike traditional stereo vision approaches that might struggle with near-field depth estimation due to larger baselines, our sensor's compact design significantly reduces blind spots in close proximity. Consequently, the sensor surface can be positioned as close as about 11mm from the imaging system, marking an advancement in compact sensor design. Calibration results are detailed in Table \ref{tab:parameters}.

\begin{table}[!t]
\vspace{0.25cm}
\caption{Stereo Calibration Results}
\label{tab:parameters}
\begin{center}
\begin{tabular}{|c|c|c|}
\hline
Stereo Units & Left & Right\\
\hline
Focal Length (pixels) & 443.50 & 387.96 \\
\hline
Baseline (pixels) & 2.02 & 2.02 \\
\hline
Reprojection Error (pixels) & 0.54 & 0.24 \\
\hline
\end{tabular}
\end{center}
\vspace{-0.25cm}
\end{table}

As illustrated in Fig. \ref{fig:stereo_process}, the depth estimation process begins by extracting image tiles from the specified stereo pair. The vertical arrangement of these units, unlike traditional setups, necessitates the transposition of tiles to align left and right images for standard stereo processing. This is followed by distortion correction to rectify any irregularities. The tiles are then converted to grayscale in preparation for applying the Semi-Global Block Matching (SGBM) algorithm \cite{sgbm}\cite{opencv}.
Subsequently, a right disparity map is generated, which is essential for applying Weighted Least Squares (WLS) filtering. WLS filtering enhances the disparity map by reducing noise and improving image quality. Finally, the refined disparity map is obtained. Any values below the minimum disparity are set to NaN, which helps handle ambiguous regions, such as repetitive or texture-less areas, where matching algorithms struggle to identify unique matches. Furthermore, any values within the marker area of the disparity map are excluded, as the depth of markers does not contribute to our study.

\subsection{Tactile Modality}

The tactile sensing process comprises three steps:

\subsubsection{Image Stitching}

As shown in Fig. \ref{fig:tactile_process}(a), the central vision unit captures eight markers, including two centrally located and six near the edges. Upon contact, there's a risk that edge markers move out of view and become untrackable by a single unit. To address this, surrounding units can be used to expand the field of view, ensuring all markers are tracked with sufficient margins for post-contact movement.

The process starts with capturing an initial image, featuring a clear white background without any contact on the sensor surface. This background simplifies blob detection and aids in determining the centers of common markers, which serve as references for defining Regions of Interest (ROIs). Selected ROIs prevent overlaps between units and exclude black and unused areas, ensuring each unit's view is unique and effectively utilized.
After identifying the ROIs, they are cropped and stitched into a single image. Upon applying force, the same cropping regions produce a stitched post-contact image, as shown in Fig. \ref{fig:tactile_process}(b). This stitched image ensures the tracking of all markers after contact.

\subsubsection{Blob Detection}

Our sensor utilizes the HSV (Hue, Saturation, Value) color space adjustment method to enhance the isolation of markers.
By fine-tuning the HSV filter values, we can mitigate specific components that could potentially impede marker detection. Subsequently, the SimpleBlobDetector from OpenCV \cite{opencv} is employed for the detection process within the masked regions, as shown in Fig. \ref{fig:tactile_process}(c). This method allows for the accurate identification of markers and the extraction of their centroid coordinates.

\subsubsection{Marker Displacement Extraction}

After the initial image capture, all markers are precisely detected for tracking. For each frame that follows, the detected blob points are compared and matched with those from the initial image, establishing correspondences based on nearest neighbors. These displacements represent contact deformation, thereby generating a 2D flow. The result of this processing is illustrated in Fig. \ref{fig:tactile_process}(d).
The 2D flow is converted into a 6x6 grid through cubic interpolation of sparse point displacements, utilizing a SciPy function \cite{scipy}. This step enables the creation of a continuous field from markers, regardless of their sparse distribution or irregular spacing.

\subsection{Simultaneous 3D Visual and Tactile Sensing}

The sensor can process both 3D visual and tactile data from a single image, owing to the compound-eye imaging system's multiple vision units, each assigned to a specific task. The dual-focus microlens array within the system enables simultaneous 3D visual and tactile sensing without requiring modality conversion. By using multi-threading on a computer with an i9-10900 CPU, the sensor achieves a sensing rate of 25 frames per second. This method significantly enhances operational efficiency and responsiveness.

\section{Experiments}

In this section, we detail two experiments to calibrate and evaluate the performance of depth estimation and force measurement.

\subsection{Depth Estimation Calibration}

We trained and evaluated the performance of our stereo depth estimation. We employed a series of performance metrics—fill rate, Z-accuracy, Root Mean Square Error (RMSE), and temporal noise—to evaluate the system's efficacy \cite{realsense}. A thorough evaluation of the performance of both the left and right stereo pairs was carried out.

\begin{figure}[!t]
\vspace{0.25cm}
  \centering
  \includegraphics[width=0.9\linewidth]{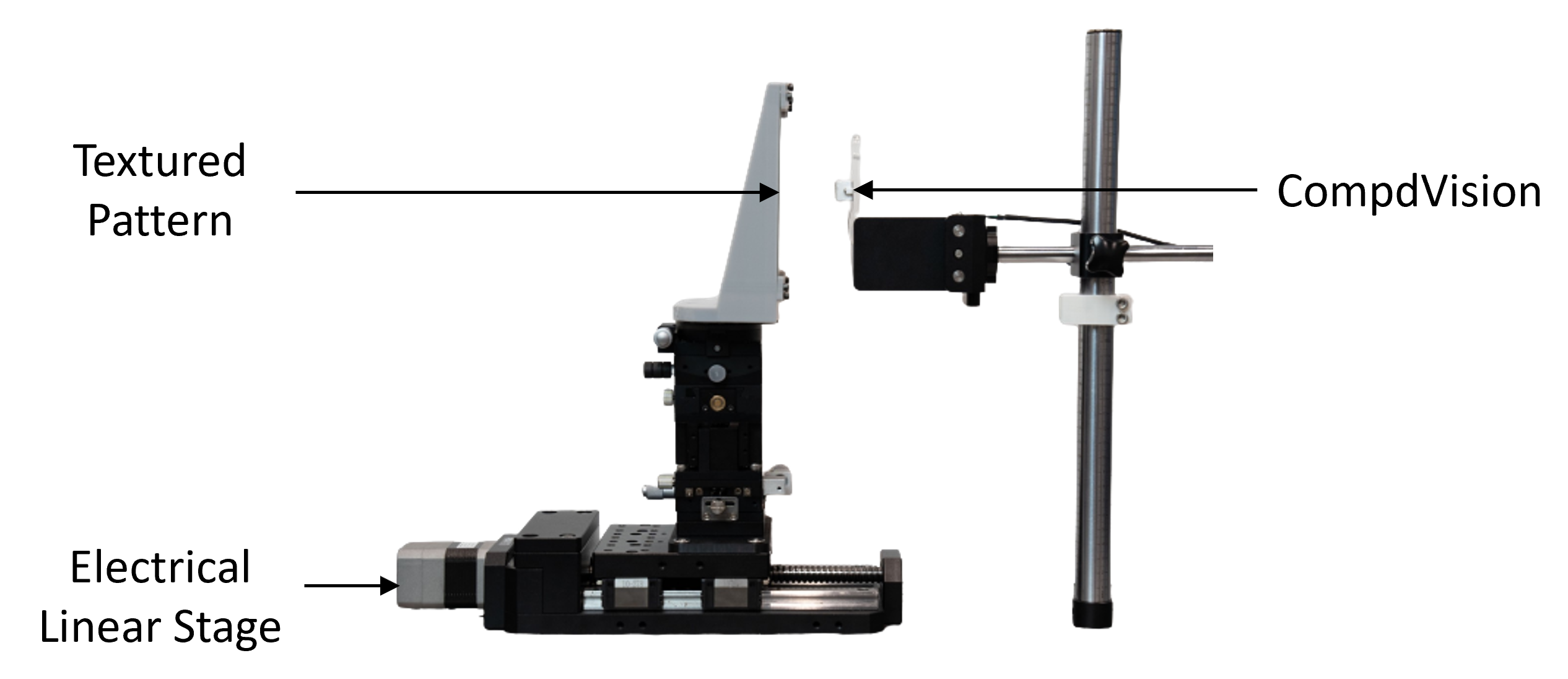}
  \caption{The experiment setup of stereo units dataset collection.}
  \label{fig:stereo_setup}
  \vspace{-0.25cm}
\end{figure}

We used a setup featuring a textured target and an electrical linear stage to generate a dataset, as illustrated in Fig. \ref{fig:stereo_setup}. The textured target served as the object for depth mapping. The ground truth for depth was established by moving the electrical linear stage, which operated within a 0 to 70 mm range from the sensor surface. For each 1 mm interval within this range, we collected 20 images. The collected dataset was divided into two subsets: 60\% for training and 40\% for evaluation.

\begin{figure*}[!t]
\vspace{0.25cm}
  \centering
  \includegraphics[width=\linewidth]{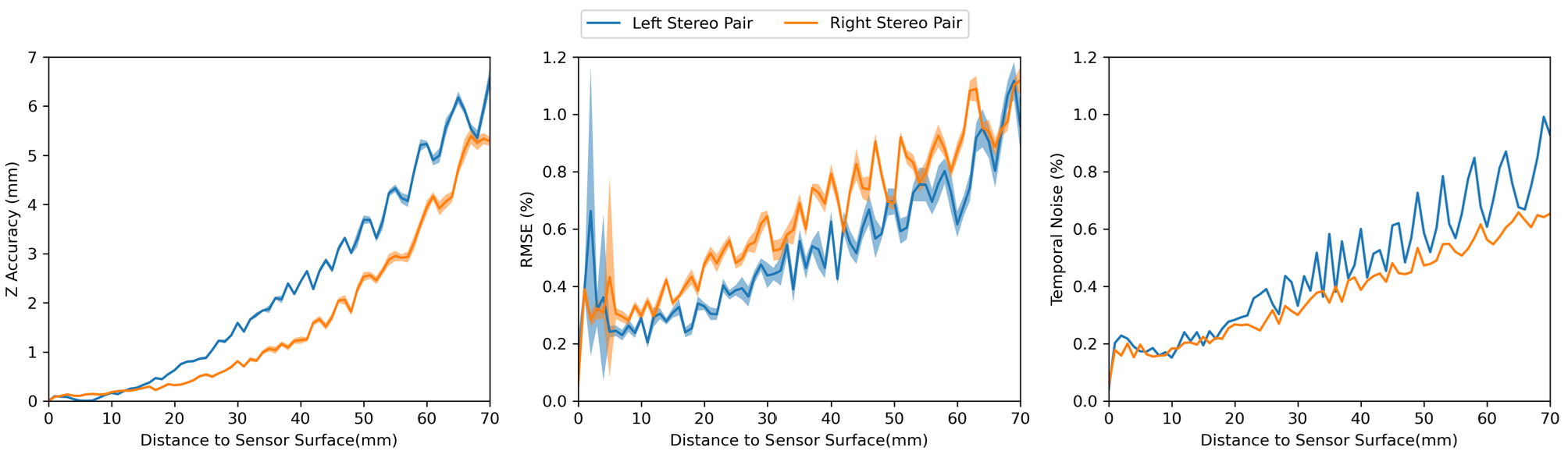}
  \vspace{-0.7cm}
  \caption{Evaluation results of depth estimation.}
  \label{fig:stereo_result}
  \vspace{-0.25cm}
\end{figure*}

For the training process, we employed the dual annealing algorithm, using the training set to train the model. This global optimization method \cite{dual_annealing} was well-suited to our complex, multi-modal cost function, which involves multiple parameters for the SGBM algorithm and the WLS filter.

The cost function was designed to optimize for two key factors: RMSE, where a lower value indicates higher depth map quality, and a constraint on the fill rate, which must exceed 90\% to ensure a reliable depth map. Mathematically, the cost function is expressed as
\begin{equation} 
Cost = 
\begin{cases} 
\sum_{i=1}^{n} RMSE_i & \text{if fill rate}  > 90\% \\
\infty & \text{otherwise}
\end{cases}
\end{equation}

We proceeded to evaluate the system's performance using the evaluation set, applying the four specific performance metrics for the final evaluation. The corresponding results are depicted in Fig. \ref{fig:stereo_result}.

\subsubsection{Fill Rate} 

The fill rate metric measures the success rate of depth calculation as a percentage of pixels in the depth map, excluding marker areas and boundary regions where disparity calculation is limited. Calculated as the ratio of pixels with valid depth to total ROI pixels, both left and right stereo pairs achieved a 100\% mean fill rate.

\subsubsection{Z-Accuracy} 

The Z-Accuracy metric evaluates the accuracy of depth estimation per pixel compared to the ground truth (GT), defined as the sum of the flange focal distance and the distance from the target to the sensor. We calculated the absolute difference between GT and the measured depth, relative to the best-fitted plane, to reduce errors from camera positioning, resulting in a depth error map.
Mathematically, the Z-Accuracy is expressed as the median value of these differences:
\begin{equation}
Z_{accuracy} = Median|Depth - GT|
\end{equation}

The evaluation results demonstrated consistent Z-Accuracy for both left and right stereo pairs, with higher accuracy observed at closer distances to the textured target. The standard deviations (STDs) are below 0.23mm.

\subsubsection{RMSE} 

The metric evaluates spatial noise in depth measurements relative to the best-fitted plane. Lower RMSE values denote superior performance and increased accuracy in depth estimation. 

Results show that mean RMSE and STD values remain below 1.2\% and 0.51\%, respectively. However, these values tend to rise when the target nearly obscure the sensor surface, primarily due to reduced light conditions and the lack of internal lighting. Despite this, the sensor's performance remains effective. This is largely due to the white sensor shell, which allows external light to penetrate, and the microlens array's low light intensity requirements \cite{lens}. Consequently, reliable depth estimation is ensured even for objects in close proximity to the sensor surface without internal lighting.

\subsubsection{Temporal Noise} 

This metric assesses the consistency of depth measurements over time. It was calculated by taking multiple depth measurements of a static scene and then computing the standard deviation of the depth values at each pixel. Lower temporal noise indicates that the system is stable and reliable in its depth estimations over time.

\subsection{Force Measurement Calibration}

We used a learning-based method to train and evaluate the estimation of tangential and normal forces. This was achieved by analyzing the displacement field of the markers, a method that has proven to be robust \cite{gelsight2}\cite{spectac}. 

\begin{figure}[!t]
    \centering
    \begin{overpic}[width=0.8\linewidth]{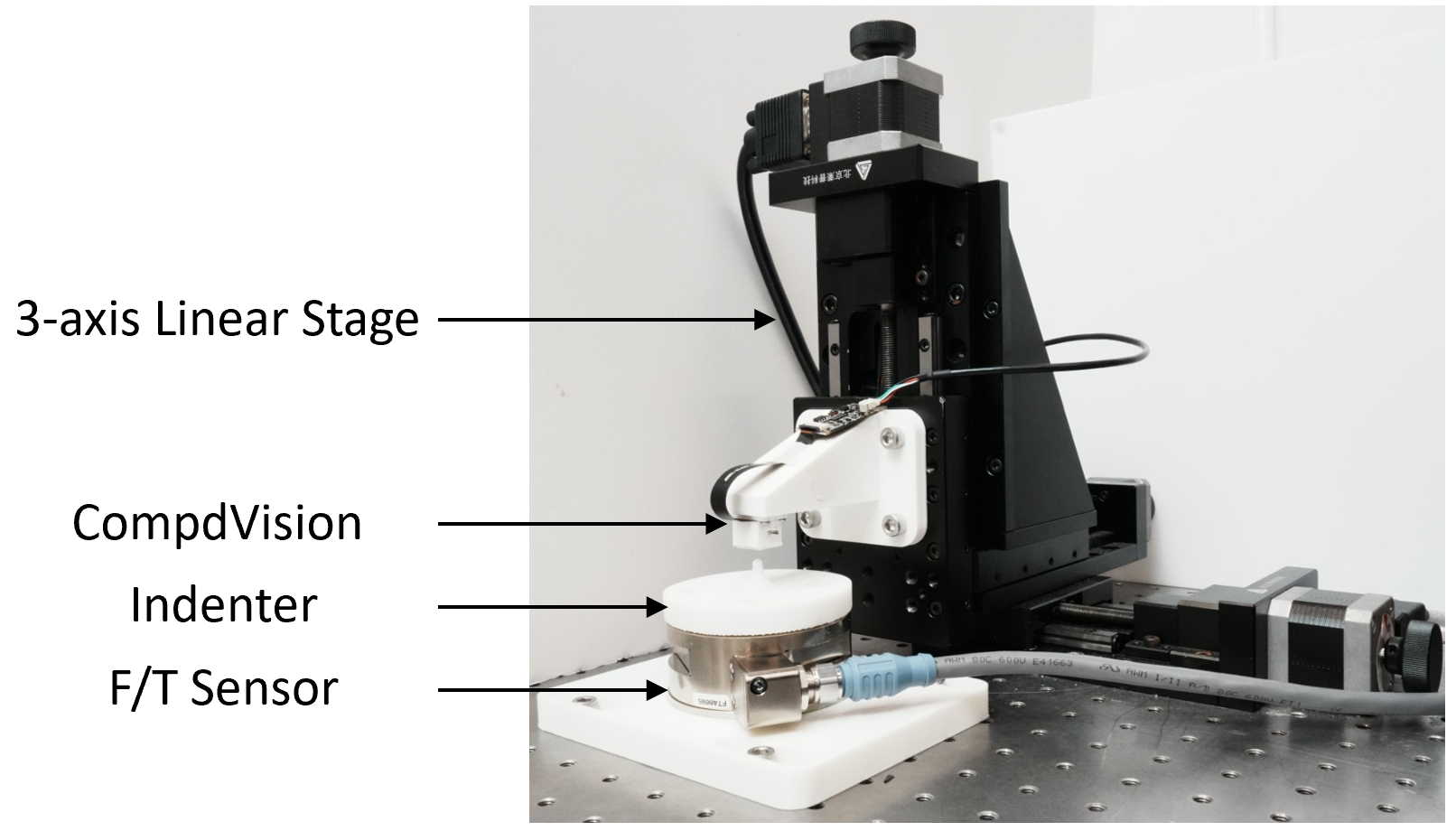}
    \end{overpic}
    \vspace{-0.25cm}
    \caption{The experiment setup of force dataset collection.}
    \label{fig:force_calibration}
\end{figure}

Fig. \ref{fig:force_calibration} illustrates our experimental setup, incorporating a 3-axis linear stage to control sensor movement. We affixed 3D-printed indenters of five different shapes (triangle, rectangle, hexagon, circular column and sphere) to the ATI Gamma Force and Torque (F/T) sensor. This setup allowed us to record and measure the ground truth force as the sensor was pressed against the various indenters with varying force amounts, directions, and locations on the sensor surface.
The sensor's interaction with indenters was controlled automatically. Indentors were pressed against the center and four surrounding points on sensor surface. For each indentation, a step size of 0.025 mm was used to incrementally increase the indentation depth up to a maximum of 4mm to introduce normal force. At each indentation step of 0.5mm, to introduce tangential force, additional movements were made up, down, left, and right by up to 3mm, each movement executed with a step size of 0.2 mm.
This procedure resulted in the introduction of both tangential and normal forces, with tangential force values ranging from -1.83N to 2.04N, and normal force values ranging from 0N to 4.85N. Consequently, a dataset comprising 16,000 measurements was created.

After collecting data, we randomly split the dataset, allocating 70\% for training and 30\% for testing. Our data processing employed a Convolutional Neural Network (CNN) model with multiple convolutional and fully connected layers. The model incorporates five convolutional layers and batch normalization for stable learning. Enhanced by max pooling and dropout for regularization, the architecture concludes with fully connected layers, leading to three output neurons for the forces $[F_{x}, F_{y}, F_{z}]$, where $F_{x}$ and $F_{y}$ denote tangential forces, and $F_{z}$ denotes the normal force.
The evaluation results using the testing dataset, as shown in Table \ref{tab:force}, indicate the RMSE and the coefficient of determination ($R^2$) for the force measurement on all three axes.
The experiments and results underscore our sensor's robust capability for contact force measurement.

\begin{table}[!t]
\caption{Evaluation Results of Force Measurement}
\label{tab:force}
\begin{center}
\begin{tabular}{|c|c|c|c|}
\hline
Force & $F_{x}$ & $F_{y}$ & $F_{z}$\\
\hline
RMSE (N) & 0.17 & 0.18 & 0.26 \\
\hline
$R^2$ & 0.84 & 0.85 & 0.95 \\
\hline
\end{tabular}
\end{center}
\vspace{-0.25cm}
\end{table}

\section{Discussion and Conclusion}

CompdVision represents a significant advancement in robotic sensing, offering a compact solution for a multi-modal sensor capable of simultaneous 3D visual and tactile sensing. Its innovative design and configuration set a new standard for multi-modal sensing in robotics.

While the sensor demonstrates promising capabilities, a thorough understanding of its limitations is crucial.  It demonstrates great depth estimation capabilities, marked by excellent Z-accuracy and RMSE, even under conditions where objects are in close proximity or obscure the sensor. However, its performance heavily depends on external lighting, a limitation that could be mitigated by incorporating internal lighting to improve low-light performance. Additionally, the sensor's reliance on black markers may pose challenges in blob detection when interacting with black objects. By applying a special coating to the markers and customizing tactile units to detect markers against diverse backgrounds, we can enhance the sensor's tactile sensing adaptability.
Future research will focus on enhancing the sensor's robustness and performance in dynamic environments and exploring its applications in robotic manipulation tasks.


\bibliographystyle{IEEEtran}
\bibliography{references}

\end{document}